%% file: main.tex
\documentclass[10pt, a4paper]{article}

\usepackage[final]{lrec2026} 

\usepackage{graphicx}
\usepackage{arydshln}
\usepackage{float}
\usepackage{listings}
\usepackage{listingsutf8}
\usepackage{placeins}
\usepackage[table,xcdraw]{xcolor}
\usepackage{amsmath}
\usepackage{multirow}
\usepackage[normalem]{ulem}
\usepackage{verbatim}
\usepackage{rotating}
\usepackage{comment}
\usepackage{dashrule}
\usepackage{booktabs}
\usepackage{tabularray}
\usepackage{siunitx}
\usepackage[most]{tcolorbox}
\title{MedInjection-FR: Exploring the Role of Native, Synthetic, and Translated Data in Biomedical Instruction Tuning}

\name{\large\textbf{Ikram Belmadani}\textsuperscript{1,2}, 
      \large\textbf{Oumaima El Khettari}\textsuperscript{2}, \\
      \large\textbf{Pacôme Constant dit Beaufils}\textsuperscript{3,4}, 
      \large\textbf{Benoit Favre}\textsuperscript{1,5},
    \large\textbf{Richard Dufour}\textsuperscript{2}}

\address{\textsuperscript{1}Aix-Marseille Univ., CNRS, LIS UMR 7020, 13000 Marseille, France, \\
\textsuperscript{2}Nantes Univ., École Centrale Nantes, CNRS, LS2N, UMR 6004, 44000 Nantes, France, \\
        \textsuperscript{3}Nantes Univ., CHU Nantes, PHU 11: Santé Publique,\\ Clinique des données, INSERM, CIC 1413, 44000 Nantes, France , \\
        \textsuperscript{4} Nantes Univ., CNRS, INSERM, L'institut du thorax, 44000 Nantes, France, \\
        \textsuperscript{5} Université Grenoble Alpes, CNRS, INRIA, Grenoble INP,\\
        \texttt{\{first.last\}@\{univ-amu.fr, univ-nantes.fr, chu-nantes.fr\}}}

\abstract{
Instruction tuning has become essential for adapting large language models (LLMs) to follow domain-specific prompts. Yet, in specialized fields such as medicine, the scarcity of high-quality French instruction data limits effective supervision. To address this gap, we introduce MedInjection-FR, a large-scale French biomedical instruction dataset comprising 571K instruction-response pairs drawn from three complementary sources: native, synthetic, and translated data. We design a controlled experimental framework to systematically assess how data provenance affects instruction tuning, using Qwen-4B-Instruct fine-tuned across seven configurations combining these sources. Results show that native data yield the strongest performance, while mixed setups, particularly native and translated, provide complementary benefits. Synthetic data alone remains less effective but contributes positively when balanced with native supervision. Evaluation on open-ended QA combines automatic metrics, LLM-as-a-judge assessment, and human expert review; although LLM-based judgments correlate best with human ratings, they show sensitivity to verbosity. These findings highlight that data authenticity and diversity jointly shape downstream adaptation and that heterogeneous supervision can mitigate the scarcity of native French medical instructions.
 \\ \newline  }

\begin{document}

\maketitleabstract

\begin{figure*}[htbp]
\centering
\includegraphics[width=\textwidth]{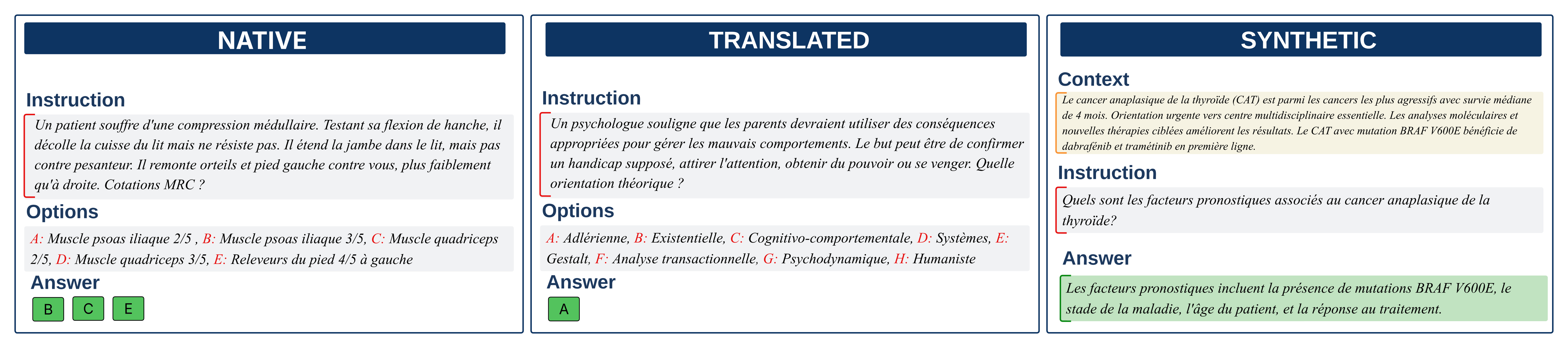}
\caption{Sample instances from the three MedInjection-FR components. Each includes the task type (MCQ, MCQU, or OEQ) and, when applicable, its supporting context. The figure illustrates the diversity of medical reasoning tasks and prompt styles across data sources.}
\label{overview}
\end{figure*}

\section{Introduction}

Supervised fine-tuning (SFT) has become the standard approach to adapting LLMs to follow instructions~\cite{shengyu2023instruction}. However, adaptation in specialized domains, such as medicine, cannot rely solely on continued pretraining over domain-specific text. While domain-adaptive pretraining (DAPT) or continual pretraining (CPT) can enrich models with in-domain vocabulary and factual knowledge, these methods do not necessarily improve, and may even erode, instruction-following behavior~\cite{christophe-etal-2024-beyond}. In practice, models exposed to extended domain pretraining sometimes ``forget'' how to respond coherently to prompts or adhere to task formats learned during earlier instruction tuning~\cite{kotha2024understanding}. These limitations make instruction-oriented supervision indispensable for aligning models with clinical reasoning and communicative expectations.

Yet, this process relies on a crucial resource: an \textbf{instruction dataset}. Such datasets are not always easy to obtain, especially in expert domains like the medical field, where creating high-quality (instruction, response) pairs requires substantial domain knowledge. Constructing these datasets often demands manual labor, either for curating the pairs themselves or for evaluating automatically generated ones to ensure factual and clinical accuracy.

Moreover, most existing biomedical instruction datasets are centered on English, with only a few exceptions for the French language, such as FrenchMedMCQA~\cite{labrak2023frenchmedmcqa} and MedQAI~\cite{bazoge2025mediqal}. This leaves a significant gap for other languages, particularly French, where linguistic, legal, and ethical limitations often constrain access to medical data. As a result, the scarcity of large, high-quality native resources raises the question of whether alternative data sources, such as automatically generated or translated instructions, can effectively complement or partially replace native supervision.

While synthetic and translated instructions can potentially increase data diversity and expand topic coverage, their true impact on downstream adaptation remains insufficiently understood. Do they actually enhance instruction-following and domain reasoning, or do they introduce stylistic and factual inconsistencies that hinder alignment? Addressing this uncertainty requires a systematic comparison of data sources under controlled conditions.

To this end, we introduce MedInjection-FR, a new French biomedical instruction dataset designed to inject medical knowledge into LLMs through diverse forms of supervision. MedInjection-FR is composed of three complementary subsets: (i) \textbf{Native} instructions derived directly from real French medical texts and resources; (ii) \textbf{Synthetic} instructions automatically generated from French biomedical abstracts and clinical cases using large generative models; and (iii) \textbf{Translated} instructions adapted from existing English biomedical instruction datasets.

This composition enables a systematic investigation of how the nature and provenance of training data influence model adaptation. We hypothesize that native data, being linguistically and culturally aligned, offer the most direct benefit for medical reasoning in French, but that combining heterogeneous sources may further enhance robustness and generalization particularly when native data are scarce.

To test this hypothesis, we conduct a series of experiments in which instruction-tuned LLMs are trained on each subset individually and on different combinations thereof. The resulting models are evaluated on multiple biomedical question answering and reasoning benchmarks. This study allows us to quantify the contribution of each data type and analyze the trade-offs between authenticity, diversity, and linguistic adaptation in instruction tuning.
Our contributions are threefold:
\begin{itemize}
    \item We release MedInjection-FR, the first open large-scale French biomedical instruction dataset, comprising 571\,436 (instruction, response) pairs that integrate native, synthetic, and translated supervision sources: \url{https://github.com/ikram28/MedInjection-FR}.
    \item We provide an experimental framework to assess the effect of data origin on the performance of instruction-tuned models.
    \item We offer empirical insights into the relative benefits of native versus augmented data for medical domain adaptation in French.
\end{itemize}

\section{Related Work}
\label{sec:related-work}

Instruction tuning has emerged as a key paradigm for aligning LLMs with human intent~\cite{lou2024large, han2025towards, liang2025aligning}. It allows the model to generalize to unseen tasks through supervised fine-tuning or instruction-response pairs. This paradigm has also been extended to the biomedical domain, with adaptations of general instruction-tuning datasets resulting in AlpaCare~\cite{zhang2023alpacare}, BioInstruct~\cite{tran2024bioinstruct}, and MedAlpaca~\cite{han2023medalpaca}, which provide diverse medical instruction-response pairs spanning a diversity of tasks. Other works, such as MedMCQA~\cite{pmlr-v174-pal22a}, MedQA~\cite{jin2021disease}, and PubMedQA~\cite{jin-etal-2019-pubmedqa}, provide large-scale benchmarks for evaluating medical reasoning and factuality, but they remain English-centric.

French biomedical datasets are scarce, and notable examples include FrenchMedMCQA~\cite{labrak2023frenchmedmcqa} and MediQAl~\cite{bazoge2025mediqal}. To overcome annotation costs and privacy constraints, recent work leverages LLMs to generate synthetic instructions ~\cite{shashidhar2025yourbench, wang-etal-2023-self-instruct, zeng-etal-2024-automatic} or translate existing datasets~\cite{kim2024instatrans}. The trade-off between native, synthetic, and translated data, however, remains an open question. MedInjection-FR addresses this by providing a French biomedical instruction dataset combining all three sources, enabling systematic study of data provenance in instruction-tuned models.

\section{Dataset Construction}
\subsection{Overview}

\begin{figure}[htbp]
\centering
\includegraphics[width=\columnwidth]{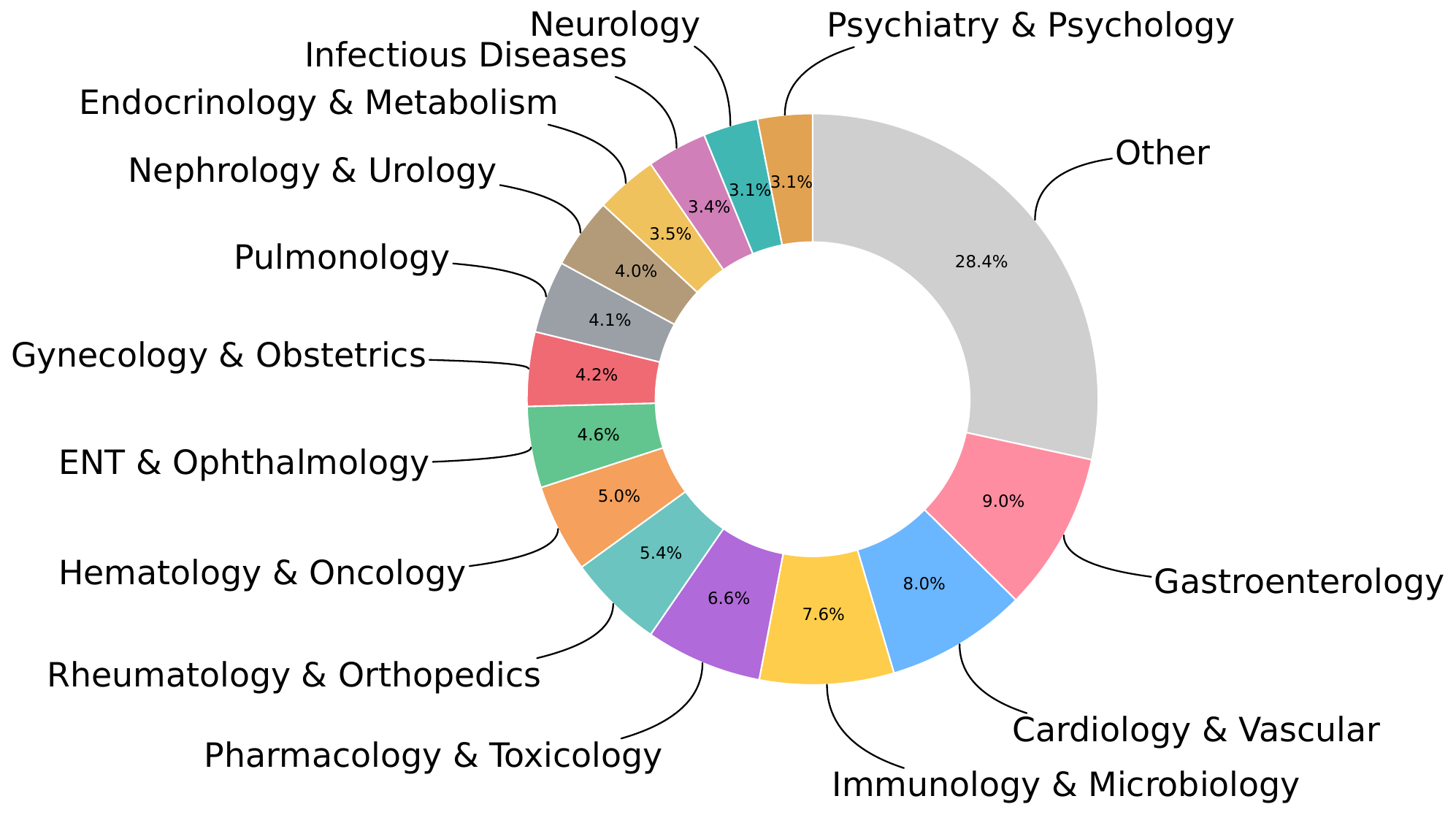}
\caption{Distribution of medical specialties in MedInjection-FR (native + synthetic components).
Proportions are normalized across fourteen major medical categories derived from specialty metadata.}
\label{specialty}
\end{figure}

The MedInjection-FR dataset comprises 571\,436 instruction-response pairs distributed across three complementary components: native, synthetic, and translated. As shown in Table~\ref{tab:medinjection-stats}, the translated subset constitutes the largest portion of the dataset (417 K pairs), followed by the native (77 K) and synthetic (76 K) subsets. Examples from each component are illustrated in Figure~\ref{overview}. Each subset is divided into training, validation, and test splits.
\input{medinjection-stats-tab}

The dataset includes three task formats: open-ended question answering (OEQ), multiple-choice question answering with multiple correct answers (MCQ), and multiple-choice question answering with a unique correct answer (MCQU). Each of these may appear either as a context-based task, in which a medical scenario is described and followed by a question-answer pair derived from that context, or a standalone instruction. Their respective distributions are reported in Table~\ref{tab:mcq-mcqu-oeq}.

Specialty metadata were available for the native and synthetic components, and grouped into fourteen broad medical categories following the standard French taxonomy. As shown in Figure~\ref{specialty}, MedInjection-FR spans a wide range of clinical and theoretical domains, with the most represented fields being gastroenterology (9\%), cardiology (8\%), immunology and microbiology (7.6\%), and pharmacology (6.6\%), supporting a general-purpose use for French medical instruction tuning. 

\input{oeq_mcq_mcqu}

\subsection{Native Data}
The native component combines curated datasets and web-scraped French medical resources to reflect authentic domain knowledge. It includes 526 question-answer pairs collected from S-Editions~\cite{S-Editions}, a French platform for educational resources for medical students, and 32\,603 items from MediQAl~\cite{bazoge2025mediqal}, which are sourced from national medical examinations covering 41 medical specialties. FrenchMedMCQA~\cite{labrak2023frenchmedmcqa} contributes 3\,105 pharmacy-focused multiple-choice questions, while two repositories by mlabonne (medical-cases-fr~\cite{mlabonne_medical_cases_fr} and medical-mcqa-fr~\cite{mlabonne_medical_mqca_fr} add 12\,194 examples originating from French medical exam databases. The FrBMedQA~\cite{kaddari2022frbmedqa} dataset complements these sources with 19\,836 questions derived from French biomedical Wikipedia articles spanning eight UMLS semantic groups (chemicals and drugs, anatomy, physiology, disorders, phenomena, procedures, genes and molecular sequences, and devices). These questions were originally formulated in a closed format and were converted into multiple-choice questions using GPT-4o-mini~\cite{hurst2024gpt} to standardize the structure across sources.

Finally, the native subset integrates parallel biomedical translation corpora derived from the WMT challenge repositories~\cite{biomedical_translation_corpora}. These bilingual data were reformulated as instruction-response pairs, where each instruction requested the French translation of an English biomedical passage.

\subsection{Translated Data}
\input{wmt}
To expand data coverage, a large collection of English biomedical instruction datasets was translated into French using Gemini 2.0 Flash~\cite{Google_Gemini2_0_Flash} and GPT-4o-mini~\cite{hurst2024gpt}. The translated component comprises 417\,674 instruction-response pairs derived from several established English resources, including MedQA~\cite{hendrycks2020measuring}, PubMedQA~\cite{jin-etal-2019-pubmedqa}, MedMCQA~\cite{pmlr-v174-pal22a}, six medical categories of MMLU~\cite{hendrycks2020measuring} , K-QA~\cite{manes-etal-2024-k}, MMLU-PRO~\citep[psychology, biology, and health]{wang2024mmlu}, and MedXpertQA~\cite{zuo2025medxpertqa}.

To assess translation quality, we evaluated the outputs on the WMT 2024 Biomedical Translation Task corpus~\cite{neves-etal-2024-findings} using BLEU and COMET metrics. As shown in Table~\ref{tab:wmt}, both models achieved performance comparable to the best WMT 2024 system, suggesting that the translated subset maintains high semantic fidelity and linguistic quality.

\subsection{Synthetic Data}
\label{subsec:syn-data}
The synthetic component was generated exclusively for the training set using GPT-4o~\cite{hurst2024gpt} as the generation model. Source contexts included clinical cases from the DEFT-2021~\cite{poulain-connes-2021-deft} and DIAMED~\cite{labrak-etal-2024-drbenchmark} datasets, and biomedical abstracts from the MORFITT~\cite{labrak-etal-2023-morfitt} corpus. From these materials, GPT-4o produced 33\,493 MCQ and 43\,013 OEQ. The complete generation prompts for each corpus and task type are reported in Appendix~\ref{app:syn-data}.

Each clinical case served as the basis for eight distinct instructional tasks: clinical note summarization, factual medical question answering, differential diagnosis support, treatment plan suggestion, laboratory result interpretation, patient instruction generation, drug interaction checking, and medical specialty classification. Abstracts were used to generate a smaller set of tasks centered on summarization, factual question answering, and specialty classification.

To assess the quality of the generated data, we employed four LLMs as evaluators: GPT-4.1-mini~\cite{OpenAI_GPT4_1}, Gemini 2.0 Flash~\cite{Google_Gemini2_0_Flash}, MedGemma-27B~\cite{sellergren2025medgemma}, and Qwen3-Next-80B-A3B-Instruct~\cite{qwen3technicalreport}. These models were selected to provide diverse perspectives on evaluation, covering variations in domain specialization (general-purpose vs. medical-specific), model size (medium vs. large), architectural design (dense vs. mixture-of-experts), and licensing (open-source vs. proprietary). 

\begin{figure}[htbp]
\centering
\includegraphics[width=\columnwidth]{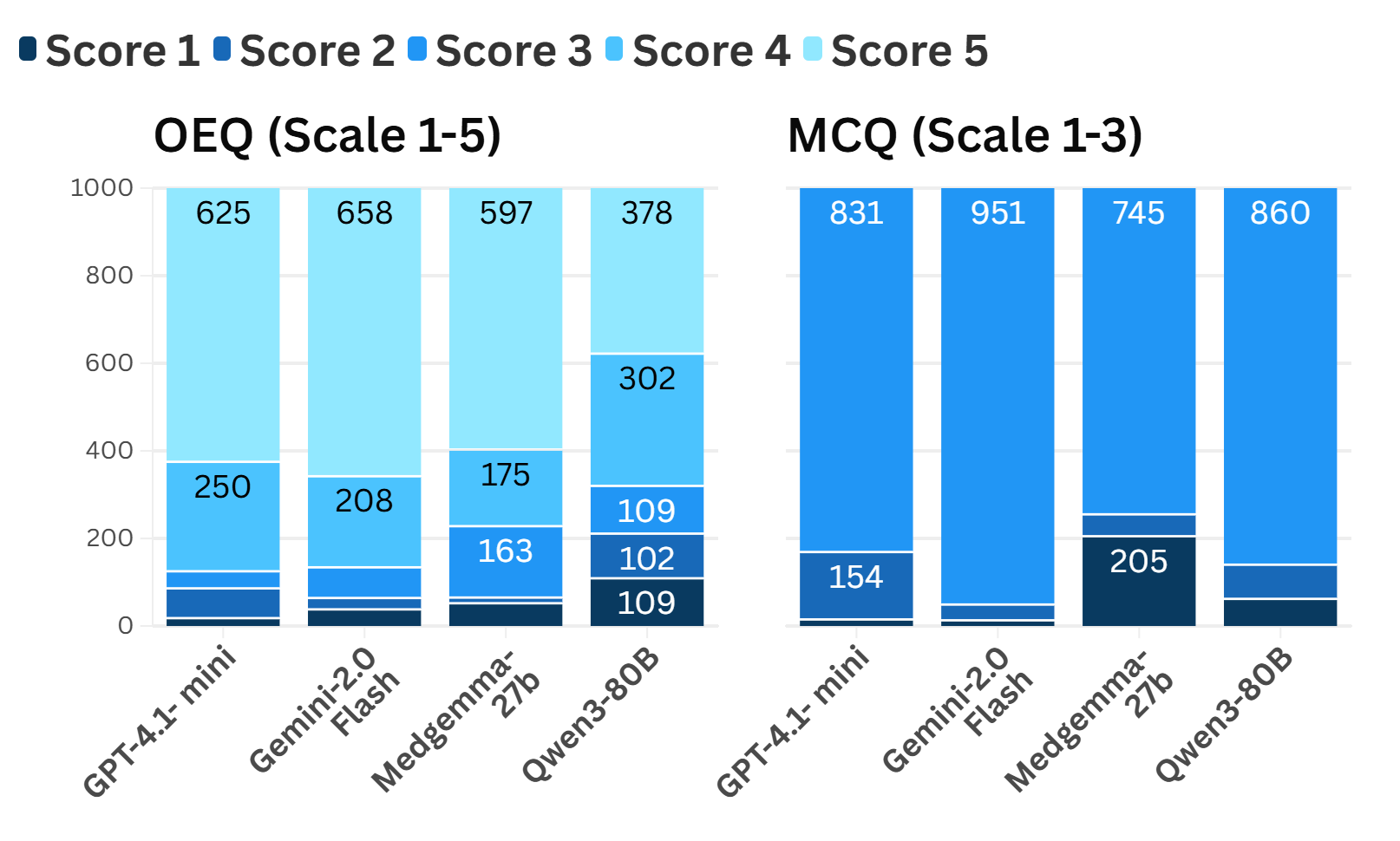}
\caption{Quality evaluation of synthetic data by four LLM judges. The left panel reports OEQ scores on a 1-5 scale, and the right panel reports MCQ scores on a 1-3 scale.}
\label{gen-quality}
\end{figure}
Each model rated the generated question-answer pairs according to predefined criteria. For MCQ data, scores ranged from 1 (irrelevant or incorrect) to 3 (accurate and contextually coherent), the prompt is specified in Figure~\ref{fig:eval-mcqa-prompt}. For OEQ data, a five-point scale was used, where 1 denoted irrelevance and 5 indicated a fully accurate and comprehensive response. The extended scale for OEQ was adopted to reflect the varying degrees of partial correctness typical of open-ended responses. The automatic evaluation prompts used for synthetic data validation are detailed in Appendix~\ref{app:syn-evaluation}.

As shown in Figure~\ref{gen-quality}, all evaluator models rated the majority of MCQ generations as high-quality (score 3), demonstrating strong contextual alignment. For OEQ, a larger spread was observed across the 3-5 range, with Gemini 2.0 Flash and GPT-4.1-mini producing the most favorable judgments overall. The results suggest that the generated subset maintains reasonable factual and contextual quality, with some variability across tasks and response types.

\section{Experimental Setup}
\label{sec:exp-setup}

To investigate how training data provenance impacts model adaptation, we conduct a controlled comparison of multiple training configurations. Specifically, we address the following research question:

\textit{How do different sources of French medical instruction data (native, synthetic, translated) impact LLM performance, and do combining these sources yield complementary benefits compared to relying on a single source?}

\subsection{Training Configurations}
To systematically evaluate this question, we constructed seven dataset configurations isolating the contribution of each data origin: NAT (native only), TRAD (translated only), SYN (synthetic only), NAT-TRAD, NAT-SYN, TRAD-SYN, and ALL (all sources combined).
Each configuration contains 33\,493 examples, corresponding to the smallest common sample size across all sources to ensure a fair comparison. The datasets consist exclusively of multiple-choice tasks, as open-ended QA data were mainly reserved for evaluation. For mixed configurations, we maintained equal proportions of examples from each contributing subset to avoid bias toward a specific source.

\begin{figure}[htbp]
\centering
\includegraphics[width=\columnwidth]{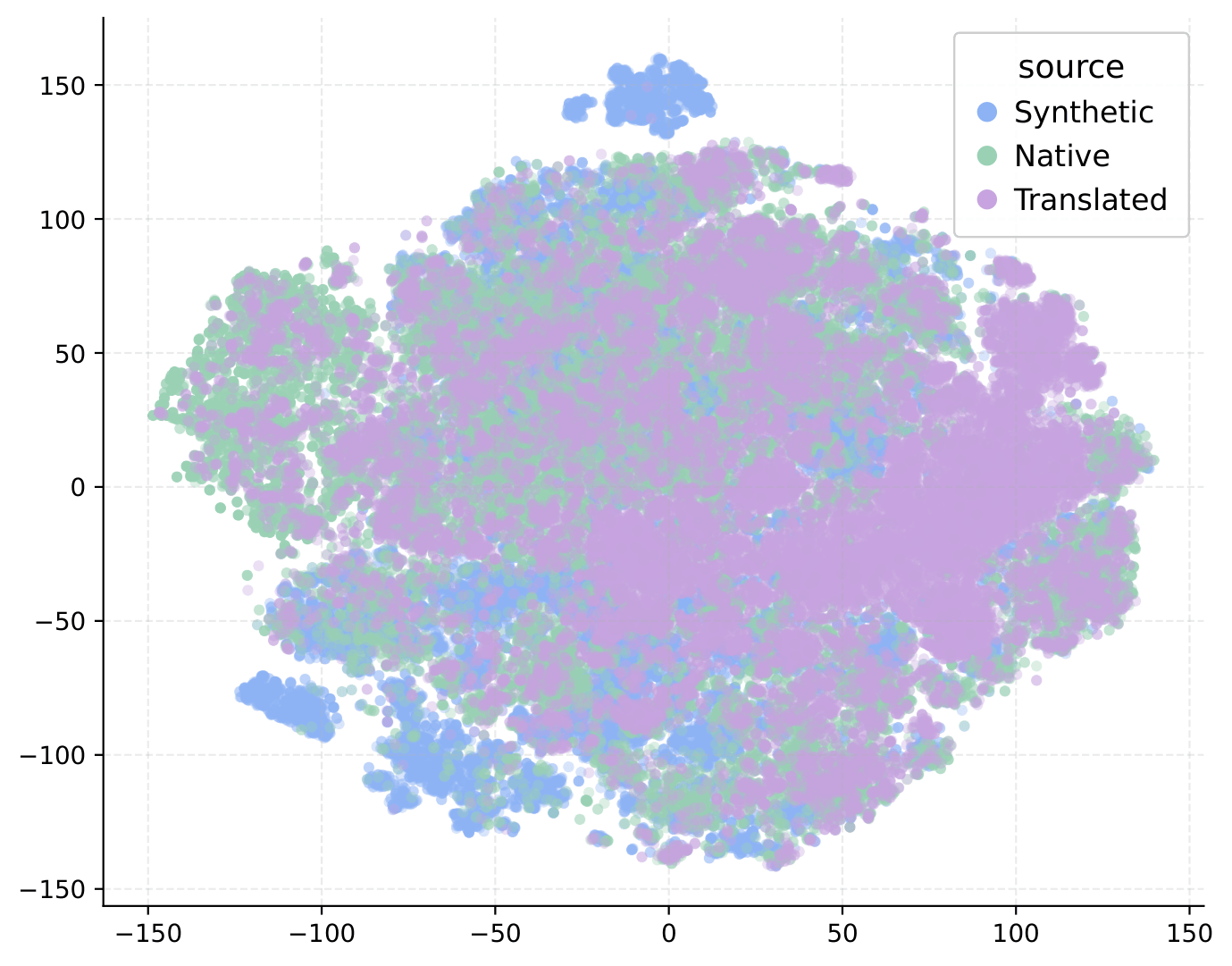}
\caption{t-SNE visualization of instruction embeddings across the three data sources (native, translated, synthetic).}
\label{fig:tsne}
\end{figure}

To examine whether the subsets differ substantially in their semantic content, we visualized instruction embeddings from each data source using t-SNE~\cite{maaten2008visualizing}. The resulting projection (Figure~\ref{fig:tsne}) provides a qualitative view of their distributional overlap. Native and translated instructions exhibit strong local mixing, suggesting similar semantic neighborhoods, while synthetic data display slightly broader dispersion, illustrating greater topical variability. Overall, the subsets appear moderately distinct yet sufficiently aligned to support joint fine-tuning without major domain shift.

This design allows for a controlled ablation study, where performance differences can be directly attributed to the inclusion or exclusion of particular data types rather than to scale effects or task imbalances.
\subsection{Base Model and Training Procedure}

We used Qwen-4B-Instruct as the base model for all experiments. This model was selected for its multilingual pretraining, compact 4B-parameter architecture, and extended context capacity that accommodates our longest inputs (\(\approx 6\,300\text{ tokens}\)) without truncation. Such coverage is particularly important for processing detailed clinical cases and multi-sentence biomedical instructions, which often require long contextual reasoning. The 4B scale offers a practical balance between representation capacity and computational cost.

We applied SFT using the DoRA~\citep[Weight Decomposed Low-Rank Adaptation]{liu2024dora} technique, an extension of LoRA~\cite{hu2022lora} that decomposes pretrained weights into magnitude and direction components. 

DoRA was chosen after preliminary comparisons with LoRA, during which it demonstrated superior stability and efficiency for small-scale biomedical fine-tuning.

Each run was trained for 10 epochs with a batch size of 12, a learning rate of $1\times 10^{-4}$, and gradient accumulation over 8 steps, using a cosine learning rate scheduler with a warmup ratio of 0.05.
For the DoRA configuration, the rank parameter was set to 16, the scaling factor (lora alpha) to 16, and the dropout probability to 0.05. These adapters were applied to the attention and feed-forward projection layers, specifically the following modules: \textit{q\_proj, k\_proj, v\_proj, o\_proj, gate\_proj, up\_proj}, and \textit{down\_proj}. All fine-tuning runs were conducted under identical training conditions, with the dataset configuration being the only varying factor.
The hyperparameters were not explicitly tuned in this study but were chosen following standard practices for adapter-based fine-tuning in instruction-tuning setups. Preliminary monitoring of training dynamics, including loss curves and gradient norms, confirmed stable optimization without signs of divergence or overfitting, supporting the suitability of these settings for our experimental framework.

\subsection{Evaluation Protocol}
\label{sec:eval-protocol}

Evaluation was carried out deterministically using greedy decoding with a temperature of~0 across all QA types (MCQ, MCQU, and OEQ). For MCQ and MCQU, performance was assessed using the Exact Match (EM) accuracy, computed under two decoding modes: (1) standard greedy decoding and (2) constrained decoding, where we constrain the token vocabulary to be one of the answer choice letters (e.g., one of [“A”, “B”, “C”, “D”] for a four-choice QA dataset) and treat the answer choice with the highest token probability as the model’s prediction. 

To improve robustness and account for potential position bias in multiple-choice evaluation~\cite{pezeshkpour-hruschka-2024-large}, we additionally performed three runs per configuration with randomized answer orderings. Reported scores correspond to the average performance across these runs. 

For MCQ tasks that include multiple correct answers, we computed the Hamming score to capture partial correctness.

The \textit{Exact Match (EM)} metric measures the proportion of predictions that exactly match the gold answer set:
\[
\text{EM} = \frac{1}{N} \sum_{i=1}^{N} [\hat{y}_i = y_i]\]
where \(N\) is the total number of questions, \(y_i\) the gold answer (or answer set), and \(\hat{y}_i\) the model prediction. 
The \textit{Hamming score} quantifies partial agreement between predicted and reference label sets:
\[
\text{Hamming Score} = \frac{1}{N} \sum_{i=1}^{N} \frac{|y_i \cap \hat{y}_i|}{|y_i \cup \hat{y}_i|}
\]

Evaluation for OEQ required a distinct approach, as the outputs are free-form text rather than discrete labels. 
To capture both linguistic similarity and factual correctness, we employed a complementary evaluation protocol combining: (1) automatic text-based metrics to quantify lexical and semantic overlap, and (2) a human and LLM-based evaluation to assess factual accuracy and clinical soundness. Sections~\ref{sec:eval-auto} and~\ref{sec:eval-human-llm} describe each component in detail.
\subsubsection{Automatic Evaluation}
\label{sec:eval-auto}
We established automatic baselines using standard text similarity metrics such as BLEU~\cite{papineni-etal-2002-bleu}, ROUGE~\cite{lin-2004-rouge}, METEOR~\cite{banerjee-lavie-2005-meteor}, and BERTScore~\cite{Zhang*2020BERTScore:}. These metrics quantify lexical and semantic overlap between model outputs and gold references. However, they remain limited in assessing factual correctness and reasoning adequacy, especially in the biomedical domain, where clinically valid answers may differ linguistically from references~\cite{yim2025morqa}.
\subsubsection{Human and LLM-as-a-Judge Evaluation}
\label{sec:eval-human-llm}
To overcome the limitations of automatic metrics, we explored an LLM-as-a-judge approach to automatically assess the quality of generated answers. 

To identify the most reliable model for this purpose, we conducted a meta-evaluation study on a sample of 100 OEQ pairs drawn from the test set. Each candidate LLM judge was prompted with the same evaluation instructions as those provided to a human medical expert. The full evaluation prompt provided to the LLM judges is reproduced in Appendix~\ref{app:eval-prompt}.

The study involved five LLMs chosen to represent a diversity of characteristics:
three general-purpose models (GPT-5~\citealp{OpenAI_GPT5}, GPT-4.1-mini~\citealp{OpenAI_GPT4_1} and Gemini 2.0 Flash~\citealp{Google_Gemini2_0_Flash}) and two biomedical models (MedGemma-27B~\citealp{sellergren2025medgemma} and HuatuoGPT-o1-72B~\citealp{chen-etal-2025-towards-medical}). 

A licensed physician independently rated the same 100 samples, assigning binary scores (1 for equivalent, 0 for non-equivalent) based on whether the generated response conveyed the essential medical information required by the reference. Minor lexical or stylistic differences were considered acceptable as long as the medical content was accurate and complete.
\input{correlation}
We then computed Pearson correlations between human ratings and the scores produced by each LLM judge and by the automatic metrics. 

As shown in Table~\ref{tab:correlation}, MedGemma-27B,the smallest of the biomedical models, exhibited the strongest alignment with expert judgments 
(\(r = 0.61,\, p = 1.74\times10^{-11}\)), outperforming both larger general models and the larger medical LLM. 
This suggests that domain specialization plays a more decisive role than model scale in capturing factual correctness. 

Traditional automatic metrics, by contrast, showed only weak to moderate correlations with human evaluation, with ROUGE-2 reaching the highest value among them (\(r \approx 0.36\)).

These findings underscore the limited reliability of lexical and embedding-based similarity measures for biomedical QA, where valid responses often employ paraphrase or alternate formulations that reduce surface overlap, while linguistically similar but clinically incorrect outputs may receive inflated scores.

\section{Results}
\label{sec:results}

In this section, we present the results obtained from the experimental configurations described in Section~\ref{sec:exp-setup}. 
For multiple-choice tasks (MCQ and MCQU), we report the scores from a single evaluation run. Results averaged over the three randomized runs (used to account for potential position bias) are provided in Appendix~\ref{app:pos-bias}. 

\input{results-mcq}

\subsection{Multiple-Choice Question Answering}
\paragraph{General trends}
Table~\ref{tab:results_mcq} summarizes the performance of all configurations on MCQ and MCQU tasks, under both greedy and constrained decoding. Overall, the results reveal consistent behavior across decoding strategies for all fine-tuned models, except for the base model. The approximately three-point gap between greedy (34.53) and constrained (37.31) decoding observed in QWEN-4B suggests that, although the base model was instruction-tuned, it was not explicitly optimized to output valid choice labels. After fine-tuning, however, this discrepancy nearly disappears, indicating that exposure to task-specific supervision helped models produce the expected answer format.
\paragraph{Task-level comparison}
Performance differences between MCQ and MCQU also reflect the relative difficulty of the tasks. Scores on MCQ are systematically lower than those on MCQU, as answering a single-correct question requires identifying one exact label, whereas MCQ allows partial correctness. This is further supported by the higher Hamming scores compared to EM values, demonstrating that the models often predict partially correct answer sets even when the EM criterion is not satisfied.
\paragraph{Effect of individual data sources}
When training on individual data sources, clear distinctions emerge. Models fine-tuned on native data (QWEN-4B-NAT) achieve the largest improvement in constrained decoding relative to the base model (40.59 EM vs. 37.31), underscoring the importance of linguistic alignment and exposure to authentic French medical discourse. In contrast, the synthetic-only configuration (QWEN-4B-SYN) yields the weakest results (29.73 EM), confirming that automatically generated data, though diverse, introduce stylistic variability and factual noise. The translated-only model (QWEN-4B-TRAD) shows a modest apparent gain under greedy decoding (36.6 EM) compared with the base model (34.53), yet this advantage disappears once decoding is constrained to valid answer labels. This pattern indicates that the observed improvement results from the decoding strategy difference rather than genuine progress in answer accuracy. 
\paragraph{Impact of mixed data configurations}
Combining multiple data types consistently improves performance compared to single-source setups. The NAT-TRAD configuration achieves the highest overall accuracy (41.29 greedy, 41.37 constrained), confirming the strong complementarity between linguistically aligned native supervision and the conceptual diversity introduced by translated English datasets. The fully mixed ALL model performs comparably (40.86 greedy, 40.97 constrained), suggesting that heterogeneous supervision enhances robustness without sacrificing linguistic coherence. The NAT-SYN configuration also yields competitive results (39.25 constrained), indicating that augmenting native data with synthetic examples can inject stylistic and task diversity while preserving factual grounding. In contrast, the TRAD-SYN configuration provides only marginal improvements over TRAD alone (36.55 vs. 36.44 constrained), suggesting limited synergy between two non-native sources. Overall, these trends highlight that native supervision plays a stabilizing role, anchoring learning when paired with noisier or cross-lingual data, and remains essential for achieving consistent gains in French medical instruction tuning.
\begin{figure}[htbp]
\centering
\includegraphics[width=\columnwidth]{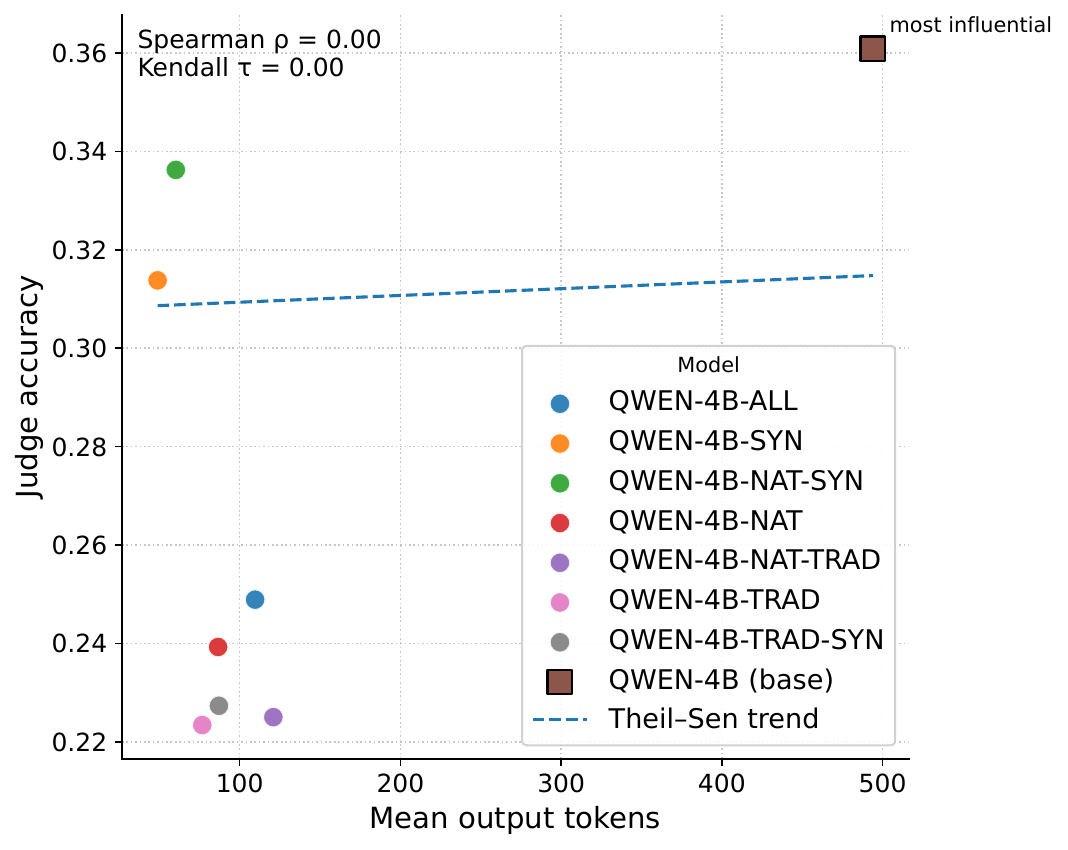}

\caption{Relationship between mean output length and LLM-judge accuracy across fine-tuned models. Each point represents one configuration, with the dashed line indicating the best-fit linear trend.}
\label{fig:verb-bias-1}
\end{figure}
\subsection{Open-Ended Question Answering}
\input{oeq-results}
Table~\ref{tab:oeq-results} presents the results for OEQ, evaluated using both automatic metrics and the LLM-as-a-judge framework. Overall, traditional text similarity metrics (BLEU, ROUGE, METEOR, BERTScore) exhibit moderate sensitivity to fine-tuning, with BLEU exhibiting the largest improvement (+20 points) and other metrics such as ROUGE-2 and BERTScore rising more modestly (+3-8 points). However, as discussed in Section~\ref{sec:eval-protocol}, these metrics are restricted to surface-level overlap.

\begin{figure}[htbp]
\centering
\includegraphics[width=\columnwidth]{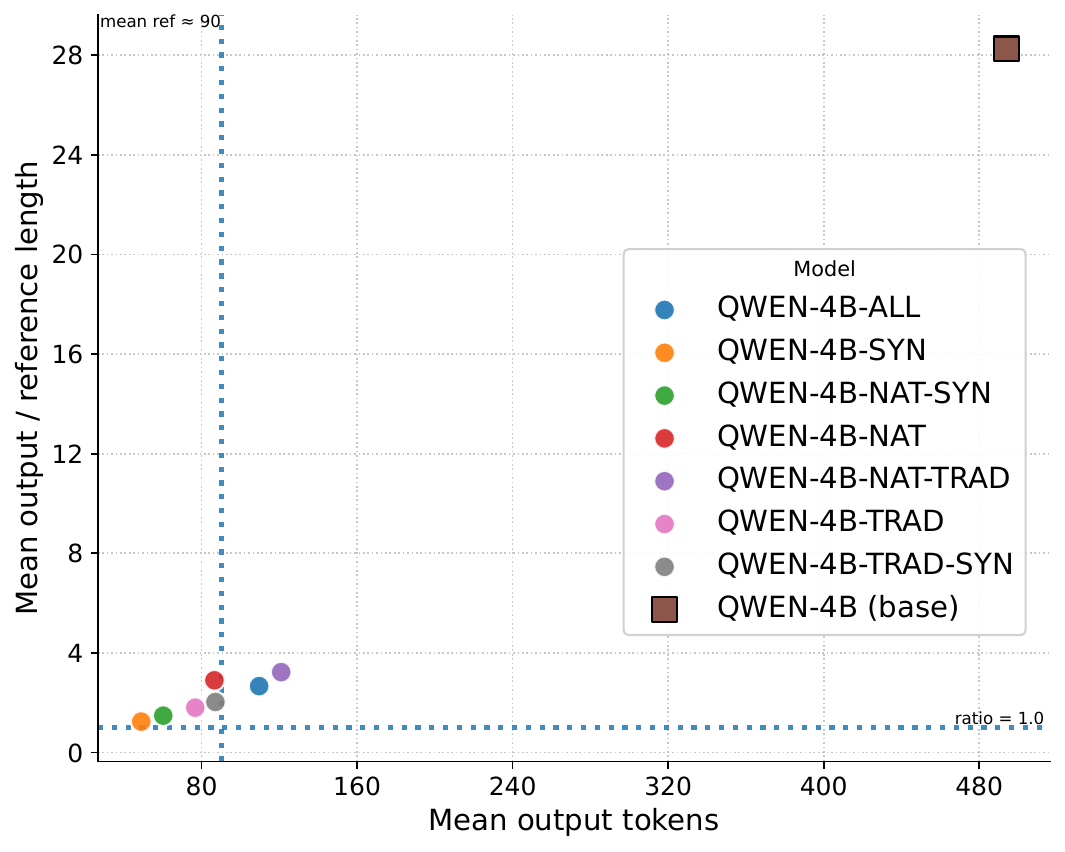}
\caption{Comparison between average output length and reference length for each model. The vertical dashed line marks the mean reference length (\(\approx 90\ tokens\)), and the horizontal line represents a perfect length ratio of 1.}
\label{fig:verb-bias-2}
\end{figure}

A notable observation from the LLM-as-a-judge evaluation is that the base model achieves the highest mean score (0.36), surpassing all fine-tuned variants. Because the base model also produces substantially longer responses, we examined whether this effect could reflect a potential verbosity bias, i.e., the tendency of evaluators to favor longer outputs~\cite{ye2025justice,Saito2023VerbosityBI}.

Figure~\ref{fig:verb-bias-1} presents the association between mean output length (log scale) and judge-assigned accuracy. Using non-parametric correlation measures, both Spearman’s $\rho$ and Kendall’s $\tau$ equal 0.00 ($p = 1$), indicating no consistent monotonic relationship across configurations. A leave-one-out sensitivity analysis further shows that the apparent association is unstable and largely influenced by a single configuration.

Figure~\ref{fig:verb-bias-2} complements this analysis by comparing the ratio between model output length and reference length. The base model stands out clearly, producing responses approximately 28 times longer than the reference answers, whereas fine-tuned models remain much closer to the target length (generally between 1 and 3.5 times the reference length). Although this visual contrast highlights a substantial verbosity difference, it does not translate into a systematic rank-based correlation across models.

Taken together, these findings suggest that while the base model’s higher score coincides with markedly longer outputs, the overall evidence for a generalized verbosity bias remains limited. The observed pattern appears configuration-specific rather than indicative of a robust relationship between response length and evaluation score.

This pattern suggests that instruction tuning on MedInjection-FR encourages models to produce more concise outputs compared to the base model. . From a linguistic perspective, models trained on native or mixed data display stronger lexical alignment and semantic adequacy, as evidenced by their higher ROUGE-2 and BERTScore values. In contrast, purely synthetic or translated data tend to underperform, likely due to minor stylistic inconsistencies or factual drift introduced during generation and translation.

\section{Discussion}
The results presented in Section~\ref{sec:results} provide insights into how the nature and provenance of instruction data affect the adaptation of French medical LLMs. Our discussion primarily draws on the MCQ and MCQU results.

\paragraph{Complementarity of data sources.}
The comparison across training configurations confirms that native French data remain the most effective for improving model adaptation, particularly under constrained decoding. 
However, when native data are limited, combining them with other sources, translated or synthetic, proves particularly valuable. Despite containing only half or one-third of the native examples, mixed configurations such as NAT-TRAD, NAT-SYN, and ALL achieve comparable or even slightly higher aggregated scores. This suggests that heterogeneous supervision may introduce complementary information that enhances generalization. When used alone, both synthetic and translated datasets remain less effective than native data, probably due to noise, translation artifacts, or stylistic inconsistencies. Nevertheless, these data types likely contribute additional variation and conceptual diversity that help models better generalize to unseen formulations. Native supervision, however, appears essential to anchor learning and mitigate noise from these less controlled sources. Such findings are especially relevant in low-resource contexts where large-scale native instruction data are difficult to obtain.

\paragraph{Insights from open-ended QA.}

The OEQ evaluation highlights several limitations of current automatic evaluation methods. Standard overlap-based metrics (e.g., BLEU, ROUGE, BERTScore) exhibit only moderate sensitivity to fine-tuning, reflecting their limited ability to capture factual or clinical correctness in biomedical settings. 

Regarding potential verbosity effects, our non-parametric analysis did not reveal a consistent monotonic relationship between response length and judge-assigned accuracy. Although the base model generates substantially longer outputs and receives higher scores, this pattern does not generalize across fine-tuned configurations, and rank-based correlation measures (Spearman’s $\rho$ and Kendall’s $\tau$) do not indicate a robust association. 

Taken together, these findings suggest two key takeaways: (1) automatic and LLM-based metrics may not reliably reflect factual precision in biomedical reasoning, and (2) evaluation of medical QA tasks benefits from domain-aware or human-in-the-loop assessment to ensure that judgments are grounded in clinical relevance rather than surface similarity or isolated configuration effects.

\section{Conclusion}
This work introduces MedInjection-FR, a comprehensive French biomedical instruction dataset designed to investigate how the provenance of supervision data affects LLM adaptation. Through a controlled series of fine-tuning experiments, we showed that while native data provide the most reliable linguistic and contextual alignment with French medical reasoning, combining them with translated or synthetic sources can achieve comparable and sometimes superior performance, even with a smaller share of native examples. This finding suggests that heterogeneous training data can represent a viable alternative when authentic French resources are scarce. Moreover, our evaluation highlights that automatic and LLM-based metrics, although most correlated with human judgments, can still be influenced by verbosity bias. Together, these results deepen our understanding of how data composition shapes instruction-tuned biomedical LLMs and underscore the need for more robust, domain-aware evaluation methods in future multilingual adaptation work.

\section{Limitations}
While the proposed study isolates the effects of data provenance under controlled conditions, several limitations remain.

First, thematic coverage and task difficulty may vary across data sources, potentially influencing performance independently of linguistic origin. Although we control for sample size and task format, fully disentangling content effects from provenance effects remains challenging.

Second, synthetic data quality was assessed using a numerical scoring scheme (Section~\ref{subsec:syn-data}) rather than qualitative or adjectival judgments (e.g., Likert-style descriptors). While numerical scores facilitate aggregation and comparison across evaluators, they may compress nuanced distinctions in answer quality. Moreover, although we report aggregate quality statistics, synthetic instances were not filtered based on these scores prior to fine-tuning. As a result, residual quality variation may have influenced downstream performance. A more detailed qualitative error analysis or selective filtering strategy could provide further insight into the relationship between generation quality and model adaptation.

Third, the evaluation relies partly on automatic and LLM-based scoring, both of which can misrepresent factual correctness or penalize concise yet accurate outputs. Although we calibrate the LLM-as-a-judge using human annotations, such approaches remain imperfect proxies for expert evaluation.

Finally, our experiments are limited to a single backbone model (Qwen-4B-Instruct). Scaling to larger architectures or models with different pretraining distributions may reveal distinct adaptation dynamics or different sensitivity to data provenance.

\section{Ethical Considerations}

This work introduces MedInjection-FR, a French biomedical instruction dataset composed of native, translated, and synthetically generated data. All source materials were drawn from publicly available biomedical datasets and educational resources. No personally identifiable health information (PHI) was used, and no patient-level private data were collected or generated.

The released dataset and fine-tuned models are provided solely for research purposes in natural language processing and machine learning.

Finally, the observed sensitivity to positional bias in multiple-choice evaluation (Appendix~\ref{app:pos-bias}) highlights broader methodological concerns in benchmarking medical LLMs. Future work should prioritize evaluation protocols that reduce unintended biases and better reflect clinically meaningful reasoning.

\section{Acknowledgements}
This work was financially supported by ANR MALADES (ANR-23-IAS1-0005). It was provided with computing HPC and storage resources by GENCI at IDRIS thanks to the grants 2025-AD011015256R1 and 2025-AD011016540 on the supercomputer Jean Zay’s H100 partition.

\section{Bibliographical References}
\label{sec:reference}
\bibliographystyle{lrec2026-natbib}

\bibliography{ref,languageresource}

\section{Appendices}
\subsection{Synthetic Data Generation}
\label{app:syn-data}
This appendix reports the exact system prompts used to generate the synthetic instruction--response pairs.
We provide separate prompts for each source corpus (MORFITT abstracts, DEFT clinical cases, and DiaMed cases)
and for each task format (OEQA and MCQA), in order to ensure full transparency and reproducibility of the
generation process. The OEQA generation prompts are provided in Figures~\ref{fig:oeqa-morfitt-prompt}--\ref{fig:oeqa-diamed-prompt},
and the MCQA generation prompts are provided in Figures~\ref{fig:mcqa-morfitt-prompt} and \ref{fig:mcqa-deft-diamed-prompt}.

\newtcolorbox{promptbox}[1]{
  colback=purple!7,
  colframe=purple!90,
  boxrule=0.6pt,
  arc=3pt,
  left=6pt, right=6pt, top=6pt, bottom=6pt,
  title=\textbf{#1},
  fonttitle=\small,
  verbatim,                  
  listing only,              
  boxsep=0pt,
  lefttitle=6pt,
  toptitle=1pt
}

\begin{figure*}[t]
\centering
\begin{promptbox}{System prompt (Morfitt Corpus)}
\begin{lstlisting}
You are a medical expert tasked with generating complex instruction-output pairs from a French biomedical abstract. The pairs should reflect hospital-relevant tasks and be sophisticated enough that only a doctor could answer them accurately, avoiding simple or direct questions (e.g., about sample size or obvious facts). 

From this abstract, generate 6 instruction-output pairs for the following 3 tasks:
1. Clinical Note Summarization (task 1): 1 pair, based strictly on the abstract.
2. Factual Medical Question Answering (task 2): 4 pairs, questions must be standalone, general medical queries inspired by the abstract’s content but phrased without any reference to the study itself; answers must reflect knowledge from the abstract.
3. Specialty Classification (task 3): 1 pair, instructing to classify the text, with the output listing the specialties provided.

Follow these instructions:
1. Format each pair as a JSON object with three fields:
   - "task": An integer (1-3) corresponding to the task number above.
   - "instruction": The instruction or question in French.
   - "output": A concise, factual answer based solely on the text (no speculation beyond what’s implied).
3. Ensure questions require clinical reasoning, synthesis, or expertise (e.g., differential diagnosis, treatment rationale), not basic recall. 
4. Return the pairs in a JSON list, e.g.:
   [
       {"task": , "instruction": "", "output": ""},
   ]
\end{lstlisting}
\end{promptbox}
\caption{System prompt used to generate OEQA synthetic instruction--response pairs from the MORFITT abstract corpus.}
\label{fig:oeqa-morfitt-prompt}
\end{figure*}

\begin{figure*}[t]
\centering
\begin{promptbox}{System prompt (DEFT Corpus)}
\begin{lstlisting}
You are a medical expert tasked with generating complex instruction-output pairs from a French clinical case report. The pairs should reflect hospital-relevant tasks and be sophisticated enough that only a doctor could answer them accurately, avoiding simple or direct questions (e.g., about age, sex, or obvious facts).
From the text, generate 15 instruction-output pairs for the following 8 tasks:
1. Clinical Note Summarization (task 1) : 2 pairs
2. Factual Medical Question Answering (task 2) : 2 pairs
3. Differential Diagnosis Support (task 3) : 2 pairs
4. Treatment Plan Suggestion (task 4): 2 pairs
5. Lab Result Interpretation (task 5): 2 pairs
6. Patient Instruction Generation (task 6): 2 pairs
7. Drug Interaction Check (task 7): 2 pairs
8. Specialty Classification (task 8): 1 pair, instructing to classify the clinical case into the relevant medical specialty (or specialties) it pertains to. Return the identified specialty or specialties in French as a list (e.g., ["Cardiologie", "Oncologie"]).
Follow these instructions:
1. Format each pair as a JSON object with three fields:
   - "task": An integer (1-8) corresponding to the task number above.
   - "instruction": The instruction or question in French.
   - "output": A concise, factual answer in French based solely on the text (no speculation beyond what’s implied).
3. Ensure questions require clinical reasoning, synthesis, or expertise (e.g., differential diagnosis, treatment rationale), not basic recall.
4. Return the pairs in a JSON list, e.g.:
   [
       {"task": , "instruction": "", "output": ""},
   ]
\end{lstlisting}
\end{promptbox}
\caption{System prompt used to generate OEQA synthetic instruction--response pairs from DEFT clinical cases.}
\label{fig:oeqa-deft-prompt}
\end{figure*}

\begin{figure*}[t]
\centering
\begin{promptbox}{System prompt (DiaMed Corpus)}
\begin{lstlisting}
You are a medical expert tasked with generating complex instruction-output pairs from a French clinical case report for fine-tuning a language model. The pairs should reflect hospital-relevant tasks and be sophisticated enough that only a doctor could answer them accurately, avoiding simple or direct questions (e.g., about age, sex, or obvious facts).
From the text, generate 16 instruction-output pairs (2 pairs per task) for the following 8 tasks:
1. Clinical Note Summarization (task 1)
2. Factual Medical Question Answering (task 2)
3. Differential Diagnosis Support (task 3)
4. Treatment Plan Suggestion (task 4)
5. Lab Result Interpretation (task 5)
6. Patient Instruction Generation (task 6)
7. Billing/ICD-10 Code Suggestion (task 7)
8. Drug Interaction Check (task 8)

Follow these instructions:
1. Format each pair as a JSON object with three fields:
   - "task": An integer (1-8) corresponding to the task number above.
   - "instruction": The instruction or question in French.
   - "output": A concise, factual answer based solely on the text (no speculation beyond what’s implied).
3. Ensure questions require clinical reasoning, synthesis, or expertise (e.g., differential diagnosis, treatment rationale), not basic recall. 
4. Return the pairs in a JSON list, e.g.:
   [
       {"task": , "instruction": "", "output": ""},
   ]
\end{lstlisting}
\end{promptbox}
\caption{System prompt used to generate OEQA synthetic instruction--response pairs from DiaMed clinical cases.}
\label{fig:oeqa-diamed-prompt}
\end{figure*}

\begin{figure*}[t]
\centering
\begin{promptbox}{System prompt (Morfitt Corpus)}
\begin{lstlisting}
You are a medical expert tasked with generating standalone multiple-choice questions (MCQAs) from a French biomedical abstract. These MCQAs will be used to fine-tune a language model for general medical knowledge reasoning.

Task:
Generate 5 MCQA pairs for Factual Medical Question Answering task.

Guidelines
- Each question must be standalone - do not reference the abstract, article, or any specific study.
- Questions must reflect general medical knowledge that can be inferred or derived from the abstract’s content.
- Focus on factual, clinically relevant knowledge: definitions, mechanisms, risk factors, complications, disease characteristics, etc.
- Avoid trivial or surface-level recall; require some synthesis or application.
- Do not copy phrases or wording directly from the abstract.
- Use French for all content.

Output format:
Return a JSON list of 5 MCQA items. Each item must follow this schema:
 [ 
   {
    "question": "",
    "options": {
      "A": "",
      "B": "",
      "C": "",
      "D": ""
    },
    "answer": <Letter(s) corresponding to the correct option choice>
  }
]
\end{lstlisting}
\end{promptbox}
\caption{System prompt used to generate MCQA items from the MORFITT abstract corpus.}
\label{fig:mcqa-morfitt-prompt}
\end{figure*}

\begin{figure*}[t]
\centering
\begin{promptbox}{System prompt (DEFT \& DiaMed Corpus)}
\begin{lstlisting}
You are a medical expert tasked with generating complex, clinically relevant MCQAs from a French clinical case report. These MCQAs will be used to fine-tune a language model and must reflect real hospital tasks.

Requirements:
- Questions must require clinical reasoning, decision-making, or synthesis of information.
- Avoid trivial questions (e.g., patient age, gender, or obvious facts stated directly).
- Design questions only a trained medical doctor could answer accurately.
- Some questions may have more than one correct answer.

Tasks:
Generate 14 MCQA pairs (2 per task) across the following 7 categories:
1. Factual Medical Question Answering - questions about pathophysiology, definitions, or clinical facts requiring reasoning.
2. Differential Diagnosis Support - identify the most probable cause given a set of symptoms.
3. Treatment Plan Suggestion - suggest the best next step or therapeutic intervention.
4. Lab Result Interpretation - interpret bloodwork, imaging, or test outcomes in context.
5. Patient Instruction Generation - what a clinician would explain or recommend to the patient.
6. Billing/ICD-10 Code Suggestion - infer the most appropriate billing or diagnostic code.
7. Drug Interaction Check - identify potential interactions or contraindications in a treatment plan.
Output format:
Return a JSON list of 14 questions. Each item should follow this schema:
[
  {
    "task": ,
    "question": "",
    "options": {
      "A": "",
      "B": "",
      "C": "",
      "D": ""
    },
    "answer": <Letter(s) corresponding to the correct option choice>
  } 
]
\end{lstlisting}
\end{promptbox}
\caption{System prompt used to generate MCQA items from DEFT and DiaMed clinical cases.}
\label{fig:mcqa-deft-diamed-prompt}
\end{figure*}

\subsection{Synthetic Data Evaluation}
\label{app:syn-evaluation}
Here we report the exact evaluation prompts used to assess the quality of the synthetic data.
We provide separate prompts for OEQA and MCQA formats, detailing the scoring criteria given to the evaluator models. The OEQA and MCQA evaluation
prompts are shown in Figures~\ref{fig:eval-oeqa-prompt} and \ref{fig:eval-mcqa-prompt}, respectively.

\begin{figure*}[t]
\centering
\begin{promptbox}{System prompt}
\begin{lstlisting}
You are a medical evaluator tasked with assessing question-answer pairs within a given context. Provide a score from 1 to 5 based on the provided score criteria.
Do not include any other opening, closing, or explanations.
Score criteria:
Score 1: The question-answer pair is completely irrelevant or incorrect given the context. The answer has major factual errors.
Score 2: The question is somewhat relevant but the answer has significant inaccuracies or lacks important details from the context.
Score 3: The question is relevant and the answer is mostly accurate but contains some minor factual errors or omissions.
Score 4: The question is clear and relevant, and the answer is accurate based on the context with only very minor omissions.
Score 5: The question is clear, relevant, and the answer is completely accurate and comprehensive based on the given context.

Remember, your score should consider both the relevance of the context to the medical domain and the accuracy of the question-answer pair. Return only the score no other text is allowed.
\end{lstlisting}
\end{promptbox}
\caption{System prompt used to automatically evaluate OEQA synthetic question--answer pairs.}
\label{fig:eval-oeqa-prompt}
\end{figure*}

\begin{figure*}[t]
\centering
\begin{promptbox}{System prompt}
\begin{lstlisting}
You are a medical evaluator tasked with assessing the correctness of multiple-choice question-answer pairs within a given context.
Provide a score from 1 to 5 based on the criteria below.
Do not include any other opening, closing, or explanations.

Score criteria:
Score 1:  The question-answer pair is completely irrelevant or incorrect given the context. The answer is incorrect.
Score 2: The question is relevant but the answer is incorrect given the context.
Score 3: The question is clear, relevant, and the answer is completely accurate based on the given context.

Return only the score, no other text is allowed.
\end{lstlisting}
\end{promptbox}
\caption{System prompt used to automatically evaluate MCQA synthetic question--answer items.}
\label{fig:eval-mcqa-prompt}
\end{figure*}

\subsection{Evaluation Prompt provided to LLM Judges}
\label{app:eval-prompt}
Figure~\ref{fig:judge-eval-prompt} presents the prompt provided to LLM judges to assess equivalence between candidate answers and the ground truth.
\begin{figure*}[t]
\centering
\begin{promptbox}{System prompt}
\begin{lstlisting}
You are a medical evaluator tasked with assessing whether a candidate answer is equivalent to the ground truth.
Assign a score strictly according to the criteria below. Do not include any explanations, comments, or extra text in your response.

Scoring criteria:
Score 0: Not equivalent
Score 1: Equivalent

Definition of equivalence:
Two answers are considered equivalent if the essential expected information is covered. Minor differences in wording, or additional or missing details, are acceptable as long as the candidate answer would be considered an acceptable response to the question.

Return only the score (0 or 1), nothing else.
\end{lstlisting}
\end{promptbox}
\caption{System prompt provided to LLM judges to score answer equivalence against the ground truth.}
\label{fig:judge-eval-prompt}
\end{figure*}

\subsection{Position Bias in MCQ Evaluation}
\label{app:pos-bias}

To assess the impact of potential position bias in multiple-choice evaluation~\cite{pezeshkpour-hruschka-2024-large}, we conducted three additional evaluation runs for each configuration, randomly permuting the answer choice ordering at each run. Reported values in Table~\ref{tab:results_mcq_pos_bias} correspond to the macro-average performance across datasets and across the three permutations, while standard deviations  quantify variability across runs.

\paragraph{General Trends}
\input{pos-bias-runs}
Compared to the single-run results reported in Table~\ref{tab:results_mcq}, overall EM scores decrease across all models and decoding modes when answer positions are randomized. 
For example, the base model’s aggregated constrained EM drops from 37.31 (single run) to 23.20 under permutation, and similar reductions are observed for all fine-tuned variants. 
This indicates that fixed answer ordering can inflate performance estimates, likely due to positional priors learned during instruction tuning.

Importantly, however, the relative ranking of configurations remains largely stable. 
Fine-tuned models continue to outperform the base model, and configurations including native supervision consistently rank among the strongest setups.

\paragraph{Task-Level Comparison}

As in the main analysis in Section~\ref{sec:results}, MCQ scores remain systematically lower than MCQU scores. 
The higher Hamming scores relative to Exact Match persist across permutations, indicating that models frequently predict partially correct answer sets even when failing the strict EM criterion. 
Randomizing answer order does not alter this structural pattern.

\paragraph{Effect of Individual Data Sources}

Under permutation, native-only fine-tuning (QWEN-4B-NAT) remains the strongest single-source configuration. 
Synthetic-only models (QWEN-4B-SYN) continue to yield the lowest EM scores across both MCQ and MCQU. 
Translated-only setups (QWEN-4B-TRAD) show intermediate performance. 
Although absolute scores decrease compared to the single-run setting, the hierarchy among individual data sources remains consistent.

\paragraph{Impact of Mixed Data Configurations}

Mixed configurations (NAT-TRAD, NAT-SYN, and ALL) continue to outperform most single-source models across randomized runs. 
NAT-TRAD and ALL remain among the top-performing configurations in aggregated EM, indicating that their advantage is not an artifact of fixed answer ordering.

\paragraph{Variance Across Runs}

Standard deviations across the three permutations are substantial, indicating that answer reordering can markedly affect absolute EM scores. 
This confirms the presence of non-negligible position sensitivity in multiple-choice evaluation. 
Importantly, however, variability remains comparable across configurations, and no model exhibits disproportionately higher instability relative to others. 
Thus, while position bias significantly impacts absolute accuracy, it does not fundamentally alter the comparative ranking of data configurations.

Overall, the permutation analysis confirms that the main findings reported in Section~\ref{sec:results} are robust to answer choice randomization at the comparative level. 
Nevertheless, the magnitude of the observed variance highlights the need for future work to systematically account for and mitigate positional artifacts in MCQ evaluation. 
Potential directions include answer-order averaging, permutation-invariant scoring protocols, or calibration strategies explicitly designed to reduce positional priors.

\end{document}

%% file: medinjection-stats-tab.tex
\begin{table}[t]
\centering
\setlength{\extrarowheight}{2pt}
\resizebox{\columnwidth}{!}{%
\def\arraystretch{1.08}
\begin{tabular}{|l|
                S[table-format=6.0]
                S[table-format=6.0]
                S[table-format=6.0] |
                S[table-format=6.0] |}
\hline
\textbf{Component} &
\multicolumn{1}{l}{\textbf{Train}} &
\multicolumn{1}{l}{\textbf{Validation}} &
\multicolumn{1}{l|}{\textbf{Test}} &
\multicolumn{1}{l|}{\textbf{Total}} \\ \hline
\textbf{Native}      & 57563 & 5055  & 14629 & 77247  \\
\textbf{Generated}   & 76506 & {--}  & {--}  & 76506  \\
\textbf{Translated}  & 366370 & 38011 & 13293 & 417674 \\
\hline 
\textbf{Total}       & 500439 & 43066 & 27931 & 571436 \\ \hline
\end{tabular}%
}
\caption{Number of instruction–response pairs in each component (native, synthetic, and translated) and their respective splits into training, validation, and test sets.}
\label{tab:medinjection-stats}
\end{table}

%% file: oeq_mcq_mcqu.tex
\begin{table}[h]
\centering
\setlength{\extrarowheight}{2pt}
\resizebox{0.8\columnwidth}{!}{%
\begin{tabular}{l
                S[table-format=6.0]
                S[table-format=6.0]
                S[table-format=6.0]}
\hline
\textbf{QA Type} & \textbf{OEQ} & \textbf{MCQ} & \textbf{MCQU} \\ \hline
\textbf{Count} & 57509 & 59592 & 454335 \\ \hline
\end{tabular}%
}
\caption{Counts of OEQ, MCQ and MCQU question types within MedInjection-FR.}
\label{tab:mcq-mcqu-oeq}
\end{table}

%% file: wmt.tex
\begin{table}
\centering
\setlength{\extrarowheight}{2pt}
\resizebox{0.9\columnwidth}{!}{%
\begin{tabular}{l
                S[table-format=2.2]   
                S[table-format=1.4]}  
\hline
\multicolumn{1}{c}{\textbf{Model}} &
\multicolumn{1}{c}{\textbf{BLEU}} &
\multicolumn{1}{c}{\textbf{COMET}} \\ \hline
\textbf{GPT-4o-mini}          & 51.01 & 0.8751 \\
\textbf{Gemini 2.0 Flash}     & 53.72 & 0.8783 \\
\textbf{WMT 2024 best system} & 53.54 & 0.8760 \\ \hline
\end{tabular}%
}
\caption{BLEU and COMET scores obtained by Gemini 2.0 Flash-001 and GPT-4o-mini on biomedical translation benchmarks, compared to the best WMT 2024 system.}
\label{tab:wmt}
\end{table}

%% file: correlation.tex
\begin{table}
\centering
\setlength{\extrarowheight}{2pt}
\resizebox{\columnwidth}{!}{%
\begin{tabular}{l l
                S[table-format=1.2]
                S[table-format=1.2e2]}
\cline{3-4}
 & & \textbf{pearson\_r} & \textbf{p\_value} \\ \hline
\multirow{5}{*}{\begin{turn}{90}\textbf{LLMs}\end{turn}}
  & \textbf{MedGemma-27B}      & \bfseries 0.61 & 1.70e-11 \\
  & \textbf{GEMINI-FLASH-2.0}  & 0.57           & 7.33e-10 \\
  & \textbf{GPT-4.1-mini}      & 0.49           & 1.80e-7  \\
  & \textbf{HuatuoGPT-o1-72B}  & 0.47           & 7.78e-7  \\
  & \textbf{GPT-5}             & 0.38           & 9.56e-5  \\ \hline
\multirow{6}{*}{\begin{turn}{90}\textbf{Metrics}\end{turn}}
  & \textbf{ROUGE-2}           & 0.36           & 2.22e-4  \\
  & \textbf{ROUGE-L}           & 0.30           & 2.39e-3  \\
  & \textbf{ROUGE-1}           & 0.29           & 3.48e-3  \\
  & \textbf{METEOR}            & 0.23           & 2.09e-2  \\
  & \textbf{BERTScore-F1}      & 0.17           & 9.49e-2  \\
  & \textbf{BLEU}              & 0.02           & 8.06e-1  \\ \hline
\end{tabular}%
}
\caption{Pearson correlation coefficients (\(r\)) and associated significance levels (\(p\)) between human expert judgments and automated evaluation metrics on 100 sampled OEQ responses.}
\label{tab:correlation}
\end{table}

%% file: results-mcq.tex
\begin{table*}
\centering
\setlength{\extrarowheight}{2pt}
\setlength{\tabcolsep}{3pt}
\small
\begin{tabular}{l
                S S S S
                S S   S S}
\cline{2-9}
 & \multicolumn{4}{c}{\textbf{MCQ}} &
   \multicolumn{2}{c}{\textbf{MCQU}} &
   \multicolumn{2}{c}{\textbf{Aggregation}} \\ \hline
\multirow{2}{*}{\textit{\textbf{Model}}} &
  \multicolumn{2}{c}{\textit{\textbf{Greedy}}} &
  \multicolumn{2}{c}{\textit{\textbf{Constrained}}} &
  \textit{\textbf{Greedy}} & \textit{\textbf{Constrained}} &
  \textit{\textbf{Greedy}} & \textit{\textbf{Constrained}} \\ \cline{2-9}
 &
  \textit{\textbf{EM}} & \textit{\textbf{Hamming}} &
  \textit{\textbf{EM}} & \textit{\textbf{Hamming}} &
  \multicolumn{2}{c}{\textit{\textbf{EM}}} &
  \multicolumn{2}{c}{\textit{\textbf{EM}}} \\ \hline
\textbf{QWEN-4B} (base) & 15.92 & 43.78 & 18.44 & 53.42 & 53.14 & 56.17* & 34.53 & 37.31 \\
\textbf{QWEN-4B-NAT} & \bfseries 27.5 & \bfseries 59.73 &
  \bfseries 27.5 & \bfseries 59.82 & 53.49 & 53.67 & 40.5 & 40.59 \\
\textbf{QWEN-4B-TRAD} & 18.56 & 53.08 & 18.0 & 52.8 &
  54.65 & 54.88 & 36.6 & 36.44 \\
\textbf{QWEN-4B-SYN} & 13.3 & 49.79 & 12.73 & 49.94 &
  46.46 & 46.73 & 29.88 & 29.73 \\
\textbf{QWEN-4B-NAT-TRAD} &
  25.65 & 59.21* & 25.65 & 59.33* &
  \bfseries 56.94 & \bfseries 57.1 & \bfseries 41.29 & \bfseries 41.37 \\
\textbf{QWEN-4B-NAT-SYN} &
  24.31 & 58.82 & 24.31 & 58.96 &
  53.96 & 54.19 & 39.14 & 39.25 \\
\textbf{QWEN-4B-TRAD-SYN} &
  18.69 & 52.98 & 18.38 & 52.95 &
  53.63 & 53.99 & 36.16 & 36.18 \\
\textbf{QWEN-4B-ALL} &
  25.95* & 58.59 & 25.98* & 58.77 &
  55.78* & 55.95 & 40.86* & 40.97* \\ \hline
\end{tabular}
\caption{Performance of all fine-tuning configurations on MCQ and MCQU under greedy and constrained decoding. Results are reported as Exact Match (EM) and Hamming scores. The aggregated columns report the average of EM scores across MCQ and MCQU formats. Since both settings correspond to MCQA and differ only in the number of correct options (single-answer vs. multiple-answer), we treat them as complementary variants of the same task. 
The aggregation therefore provides a compact summary of overall MCQA performance. \textbf{Bold} values indicate the best-performing models, while asterisks (*) denote the second-best results within each column.}
\label{tab:results_mcq}
\end{table*}

%% file: oeq-results.tex
\begin{table*}
\centering
\setlength{\extrarowheight}{2pt}
\setlength{\tabcolsep}{3pt}
\small
\begin{tabular}{
  l
  S 
  S 
  S 
  S 
  S 
}
\toprule
\multicolumn{1}{c}{\textbf{\textit{Model}}} &
\multicolumn{5}{c}{\textbf{OEQ}} \\
\cmidrule(lr){2-6}
& {\textbf{\textit{BLEU}}} &
  {\textbf{\textit{ROUGE-2}}} &
  {\textbf{\textit{METEOR}}} &
  {\textbf{\textit{BERTScore\_F1}}} &
  {\textbf{\textit{LLM-as-a-Judge}}} \\
\midrule
\textbf{QWEN-4B} (base)            & 12.08  & 20.82  & \bfseries 30.78 & 27.22  & \bfseries 0.36  \\
\textbf{QWEN-4B-NAT}         & 31.85*  & 22.67  & 25.58          & 33.44  & 0.24           \\
\textbf{QWEN-4B-TRAD}        & 29.48  & 21.42  & 26.49          & 27.39  & 0.22           \\
\textbf{QWEN-4B-SYN}         & 26.73  & 20.81  & 24.79          & 27.54  & 0.31*           \\
\textbf{QWEN-4B-NAT-TRAD}    & \bfseries 32.09  & 23.05*  & 25.90          & 35.60*  & 0.23           \\
\textbf{QWEN-4B-NAT-SYN}     & 30.63  & 22.36  & 23.73          & 21.65  & 0.34           \\
\textbf{QWEN-4B-TRAD-SYN}    & 28.07  & 21.94  & 25.77          & 34.98  & 0.23           \\
\textbf{QWEN-ALL}            & 31.74  & \bfseries 24.74  & 29.36*          & \bfseries 39.67  & 0.25           \\
\bottomrule
\end{tabular}
\caption{Results on OEQ evaluated with automatic text similarity metrics and the LLM-as-a-judge (MedGemma-27B) framework. \textbf{Bold} numbers mark the best results, and asterisks (*) indicate the second-best within each column.}
\label{tab:oeq-results}
\end{table*}

%% file: pos-bias-runs.tex

\begin{table*}[ht]
\centering
\setlength{\extrarowheight}{2pt}
\setlength{\tabcolsep}{2pt}
\tiny
\begin{tabular}{l
                S S S S S S S S}
\cline{2-9}
 & \multicolumn{4}{c}{\textbf{QCM}} &
   \multicolumn{2}{c}{\textbf{QCMU}} &
   \multicolumn{2}{c}{\textbf{Aggregation}} \\ \hline
\multirow{2}{*}{\textit{\textbf{Model}}} &
  \multicolumn{2}{c}{\textit{\textbf{Greedy}}} &
  \multicolumn{2}{c}{\textit{\textbf{Constrained}}} &
  \textit{\textbf{Greedy}} & \textit{\textbf{Constrained}} &
  \textit{\textbf{Greedy}} & \textit{\textbf{Constrained}} \\ \cline{2-9}
 &
  \textit{\textbf{EM}} & \textit{\textbf{Hamming}} &
  \textit{\textbf{EM}} & \textit{\textbf{Hamming}} &
  \multicolumn{2}{c}{\textit{\textbf{EM}}} &
  \multicolumn{2}{c}{\textit{\textbf{EM}}} \\ \hline
\textbf{QWEN-4B} (base)      & 10.76$^{4.43}$ & 37.09$^{5.84}$ & 12.72$^{5.10}$ & 45.10$^{7.25}$ & 30.98$^{19.18}$ & 33.68*$^{19.64}$ & 20.87$^{13.92}$ & 23.20$^{14.35}$ \\
\textbf{QWEN-4B-NAT}         & \bfseries 18.27$^{7.95}$ & \bfseries 49.20$^{9.27}$ & \bfseries 18.27$^{7.95}$ & \bfseries 49.23$^{9.32}$ &  32.80$^{18.03}$ &  32.86$^{18.14}$ & \bfseries 25.53$^{13.93}$ & \bfseries 25.56$^{14.00}$ \\
\textbf{QWEN-4B-TRAD}        & 12.48$^{5.63}$ & 45.41$^{6.83}$ & 11.91$^{5.56}$ & 44.96$^{6.86}$ & 33.45$^{18.54}$ & 33.54$^{18.70}$ & 22.97$^{13.70}$ & 22.73$^{13.79}$ \\
\textbf{QWEN-4B-SYN}         &  9.90$^{2.88}$ & 44.47$^{4.55}$ &  9.29$^{2.91}$ & 44.62$^{4.63}$ & 31.18$^{13.28}$ & 31.27$^{13.43}$ & 20.54$^{9.61}$  & 20.28$^{9.72}$ \\
\textbf{QWEN-4B-NAT-TRAD}    & 16.67$^{7.78}$ &  49.08*$^{8.78}$ &  16.67$^{7.78}$ &  49.14*$^{8.83}$ & \bfseries 33.80$^{20.04}$ & \bfseries 33.86$^{20.13}$ &  25.23$^{15.20}$ &  25.27$^{15.26}$ \\
\textbf{QWEN-4B-NAT-SYN}     & 15.81$^{7.48}$ & 48.44$^{8.92}$ & 15.81$^{7.48}$ & 48.49$^{9.00}$ & 32.70$^{18.37}$ & 32.80$^{18.53}$ & 24.26$^{14.03}$ & 24.30$^{14.13}$ \\
\textbf{QWEN-4B-TRAD-SYN}    & 12.38$^{5.47}$ & 45.57$^{6.42}$ & 12.01$^{5.52}$ & 45.45$^{6.50}$ & 33.18$^{17.71}$ & 33.30$^{17.92}$ & 22.78$^{13.11}$ & 22.65$^{13.26}$ \\
\textbf{QWEN-4B-ALL}         & 16.96*$^{7.56}$ & 48.84$^{8.59}$ & 16.98*$^{7.58}$ & 48.90$^{8.69}$ & 33.57*$^{19.15}$ & 33.63$^{19.25}$ & 25.27*$^{14.56}$ & 25.30*$^{14.63}$ \\ \hline
\end{tabular}
\caption{Performance of all configurations on QCM and QCMU under three answer-order permutations to assess positional bias. Results are macro-averaged across datasets and reported as Exact Match (EM) and Hamming scores. Standard deviations (superscript) correspond to the sample standard deviation across the three answer-order runs, reflecting sensitivity to option reordering. Aggregated columns report the average EM across QCM and QCMU. Bold values indicate the best-performing models, and asterisks (*) indicate the second-best within each column.}
\label{tab:results_mcq_pos_bias}
\end{table*}